\def\year{2022}\relax
\documentclass[letterpaper]{article} 
\usepackage{aaai22}  
\usepackage{times}  
\usepackage{helvet}  
\usepackage{courier}  
\usepackage[hyphens]{url}  
\usepackage{graphicx} 
\usepackage{natbib}  
\usepackage{caption} 
\DeclareCaptionStyle{ruled}{labelfont=normalfont,labelsep=colon,strut=off} 
\usepackage{algorithm}
\usepackage{algorithmic}
\usepackage{newfloat}
\usepackage{listings}
\floatstyle{ruled}
\newfloat{listing}{tb}{lst}{}
\floatname{listing}{Listing}

\usepackage{tabularx}
\usepackage{makecell}
\usepackage{dsfont}
\usepackage{amsmath}
\usepackage{amsfonts}
\usepackage{amssymb}
\usepackage{amsthm}
\usepackage{bbm}
\usepackage{multirow}
\usepackage{booktabs}
\usepackage{array}
\usepackage{pifont}
\usepackage{arydshln}
\usepackage{subfigure}
\usepackage{bm}

\newcommand{\algfull}{Adaptive In-Out-aware Learning (AIOL) }
\newcommand{\x}{\bm{x}}
\newcommand{\xin}{\bm{x}^{in}}
\newcommand{\xout}{\bm{x}^{out}}
\newcommand{\y}{\bm{y}}

\newcommand{\Uin}{U^{in}}
\newcommand{\Uout}{U^{out}}
\newcommand{\Utin}{U_t^{in}}
\newcommand{\Utout}{U_t^{out}}
\newcommand{\thin}{\tau^{in}}
\newcommand{\thout}{\tau^{out}}
\newcommand{\thtin}{\tau_t^{in}}
\newcommand{\thtout}{\tau_t^{out}}
\newcommand{\score}[1]{C_\theta({#1})}
\newcommand{\ent}[1]{\mathrm{H}\left({#1} \right)}
\newcommand{\s}{\;/\;}

\newsavebox\CBox
\def\textBF#1{\sbox\CBox{#1}\resizebox{\wd\CBox}{\ht\CBox}{\textbf{#1}}}

\newtheorem{theorem}{Theorem}

\newtheorem*{remark}{Remark}

\title{Exploiting Mixed Unlabeled Data for Detecting Samples of \\ Seen and Unseen Out-of-Distribution Classes}
\author{
    Yi-Xuan Sun, Wei Wang\footnote{Corresponding author.}\\
}
\affiliations{
    National Key Laboratory for Novel Software Technology\\
    Nanjing University, Nanjing 210023, China\\
    \{sunyixuan, wangw\}@lamda.nju.edu.cn
}

\begin{document}

\maketitle

\begin{abstract}
Out-of-Distribution (OOD) detection is essential in real-world applications, which has attracted increasing attention in recent years. However, most existing OOD detection methods require many labeled In-Distribution (ID) data, causing a heavy labeling cost. In this paper, we focus on the more realistic scenario, where limited labeled data and abundant unlabeled data are available, and these unlabeled data are mixed with ID and OOD samples. We propose the \algfull method, in which we employ the appropriate temperature to adaptively select potential ID and OOD samples from the mixed unlabeled data and consider the entropy over them for OOD detection. Moreover, since the test data in realistic applications may contain OOD samples whose classes are not in the mixed unlabeled data (we call them unseen OOD classes), data augmentation techniques are brought into the method to further improve the performance. The experiments are conducted on various benchmark datasets, which demonstrate the superiority of our method.
\end{abstract}

\section{Introduction}

Deep neural networks (DNNs) have achieved great success in various applications, but the success heavily relies on the assumption that the training and test data are drawn from the same distribution. In realistic scenarios, however, some Out-of-Distribution (OOD) samples may lead DNNs to make completely incorrect predictions, which is essentially harmful in many real-world applications, e.g., autonomous driving or medical diagnosis. Therefore, it is demanded that the trained model can at least correctly detect these OOD samples during the inference process, and then human intervention can be involved to deal with them.

Recently, \citet{hendrycks2016baseline} considered the OOD detection problem and proposed a baseline method with the output confidence, i.e., the maximum softmax probability. The method is based on the observation that In-Distribution (ID) samples tend to have higher output confidence than OOD samples. Some other methods \cite{liang2017enhancing,lee2018simple,sastry2020gram,Hsu2020generalized} made further improvements with some post-hoc techniques. However, these methods need many labeled ID data for training, causing a heavy labeling cost. 

To mitigate the labeling overload, the advanced works tried to utilize abundant unlabeled data. Some methods \cite{hendrycks2019using,tack2020csi,Sehwag2021ssd} employed self-supervised learning on the pure unlabeled ID data, while some other methods \cite{hendrycks2019deep, liu2020energe} were proposed to exploit the pure unlabeled OOD data. However, these methods require that the unlabeled data must be pure ID or OOD, which is hardly met in realistic applications. Recently, a few works \cite{Chen2020uasd,yu2020multi,guo2020safe} attempted to utilize unlabeled data that consist of ID and OOD samples. But these works were not developed for OOD detection, and there was no OOD sample in the test data during the inference process.

\begin{figure}[!t]
    \centering
    \includegraphics[width=0.45\textwidth]{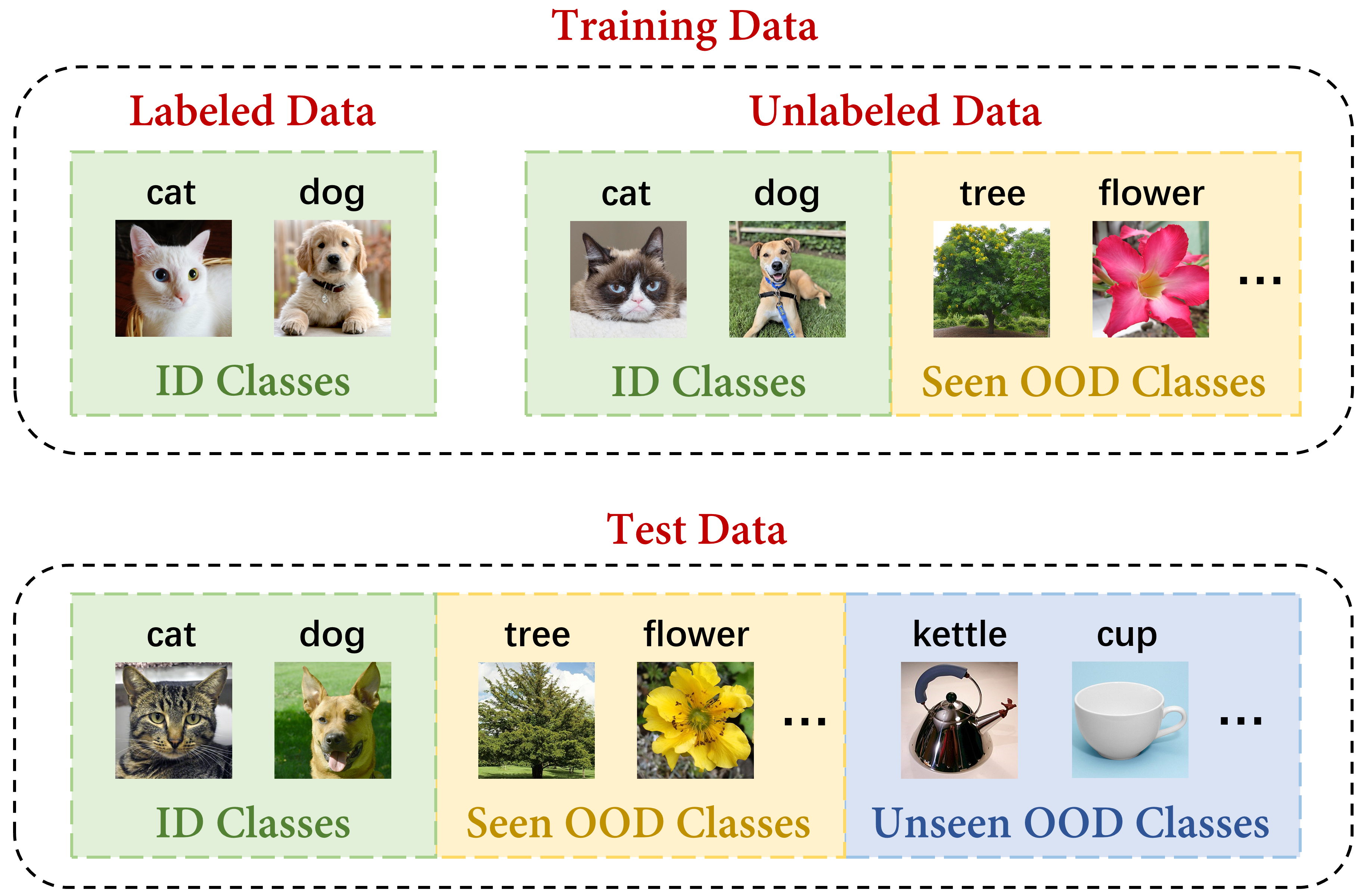}
    \caption{OOD detection with labeled and mixed unlabeled data. Unlabeled data are mixed with samples of ID classes (\textit{cat} and \textit{dog}) and seen OOD classes (\textit{tree} and \textit{flower}). Test data contain samples of ID classes (\textit{cat} and \textit{dog}), seen OOD classes (\textit{tree} and \textit{flower}) and unseen OOD classes (\textit{kettle} and \textit{cup} which are not in the mixed unlabeled data).}
    \label{fig:setting_explain}
\end{figure}

For the OOD detection problem, the unlabeled data can be mixed with ID and OOD samples. During the inference process, the test data may contain OOD samples whose classes are in the mixed unlabeled data (we call them seen OOD classes), as well as OOD samples whose classes are not in the mixed unlabeled data (we call them unseen OOD classes). The goal is to train a model which can classify ID samples and detect all OOD samples. This OOD detection problem for the image recognition task is summarized in Figure~\ref{fig:setting_explain}. A similar task exists in the medical diagnosis of lung diseases, where a model is trained with \textit{Computerized Tomography} (CT) images of the lung. Due to the heavy labeling cost, only limited labeled images (three ID classes include \textit{normal}, \textit{pneumonia}, and \textit{asthma}) and abundant unlabeled ones are available for the diagnosis of lung diseases. The images of other diseases, e.g., \textit{lung cancer}, can also be collected as unlabeled data. During the diagnosis process, the trained model should not only detect samples of seen OOD classes, e.g., \textit{lung cancer}, but also detect samples of unseen OOD classes, e.g., emerging disease \textit{COVID-19}. It is desirable to develop new methods for this realistic OOD detection problem.

In this paper, we focus on the problem discussed above and propose the \algfull method, which can utilize limited labeled data and abundant mixed unlabeled data for OOD detection. Specifically, we employ the appropriate temperature and learn the mixture probabilistic model to adaptively select the ID and OOD samples from the mixed unlabeled data. Then, the entropy over the selected samples is considered to make them more distinguishable. Moreover, since the test data may contain samples of unseen OOD classes, data augmentation techniques are brought into the method to enhance the model's generalization capability. We verify the effectiveness of our method on various benchmark datasets, and the results show that our method outperforms the compared methods.

\section{Related Works}
Out-of-Distribution (OOD) detection aims to detect OOD samples during the inference process. A baseline method \cite{hendrycks2016baseline} was proposed for detecting OOD samples with the output confidence. Later, some methods \cite{liang2017enhancing,lee2018simple,sastry2020gram,Hsu2020generalized} built the advanced detectors in a post-hoc manner. For instance, \citet{liang2017enhancing} combined temperature scaling and input preprocessing to achieve better detection performance. Instead of using the output confidence, \citet{lee2018simple} utilized the Mahalanobis distance between the test samples’ feature representations and the train samples'. However, these methods require many labeled ID samples for training.

There were some methods that focused on how to exploit unlabeled data for OOD detection. \citet{hendrycks2019deep} enforced the model to produce the low-confidence output on the pure unlabeled OOD data. Some other works \cite{golan2018geometric,hendrycks2019using,winkens2020contrastive,tack2020csi,Sehwag2021ssd} found that self-supervised learning on the pure unlabeled ID data could improve the detection performance. For instance, \citet{Sehwag2021ssd} combined contrastive learning \cite{chen2020simclr} and Mahalanobis distance for OOD detection. There was also a line of works \cite{gen0,gen1,gen2} which employed deep generative models on the pure unlabeled ID data. However, all these methods require that the unlabeled data must be pure ID or OOD, which is hardly met in realistic applications.

Recently, some methods \cite{Chen2020uasd,yu2020multi,guo2020safe} considered the class distribution mismatch between labeled and unlabeled data. The mismatched samples in the unlabeled data can be regarded as OOD samples. \citet{Chen2020uasd} filtered out OOD samples in the unlabeled data with a confidence threshold and trained the model on the remaining data only. \citet{yu2020multi} proposed a joint optimization framework to classify ID samples and filter out OOD samples concurrently. \citet{guo2020safe} employed the bi-level optimization to weaken the weights of OOD samples. But these methods were developed for ID classification and there was no OOD sample during the inference process. Another work \cite{Yu2019mcd} tried to utilize mixed unlabeled data for OOD detection. It encouraged two classifiers to maximally disagree on the mixed unlabeled data. However, each unlabeled sample was treated equally, hence the model still needed many labeled samples to distinguish between ID and OOD samples.

Semi-Supervised Learning (SSL) methods \cite{grandvalet2004semi,lee2013pseudo,berthelot2019mixmatch,xie2019unsupervised,sohn2020fixmatch} were also developed for utilizing limited labeled data and abundant unlabeled data. These methods focused on the classification performance and usually ignored the existence of OOD samples.

\section{Method}
In this paper, we consider OOD detection with labeled and unlabeled data. Let $L = \big\{ ( \x_1,y_1 ), \cdots ,( \x_n,y_n ) \big\}$ be labeled data, and $U = \{ \x_1, \cdots ,\x_m \}$ be unlabeled data. Here, $\x \in \mathcal{X} \subseteq \mathbb{R}^d$, $y \in \mathcal{Y} = \{ 1, \cdots, K \}$, $d$ is the number of the input dimension, $K$ is the number of classes in the labeled data, and $m \gg n$. For the OOD detection problem, the true class of the sample in $U$ may not belong to $\mathcal{Y} = \{ 1, \cdots, K \}$. This kind of sample is called OOD sample and is generally denoted as $\xout$, while the sample whose true class belongs to $\mathcal{Y} = \{ 1, \cdots, K \}$ is denoted as $\xin$. In this way, the labeled data can be rewritten as $L = \big\{ ( \xin_1,y_1 ), \cdots ,( \xin_n,y_n ) \big\}$. Let the ID part of $U$ be $\Uin = \{ \xin_1, \cdots , \xin_{m_1} \}$, and the OOD part of $U$ be $\Uout = \{ \xout_1, \cdots , \xout_{m_2} \}$, hence $U = \Uin \cup \Uout$. The goal of OOD detection is to learn a model $f^*$ with $L$ and $U$, i.e., $f^*(\xin)=y$ ($y \in \mathcal{Y}$ is the true class of $\xin$) for $\xin$ in the test data, while $f^*(\xout)=\bot$ ($\bot$ means the true class of $\xout$ does not belong to $\mathcal{Y}$) for $\xout$ in the test data. In realistic applications, the true class of $\xout$ in the test data may not be included in the mixed unlabeled data $U$, which is referred to as the unseen OOD class shown in Figure~\ref{fig:setting_explain}. 

We try to learn a deep neural network $f_{\theta}$ with parameters $\theta$ for OOD detection. Intuitively, we could utilize the labeled data $L$ and the mixed unlabeled data $U$ to obtain a basic model. For the labeled data $L$, we use the supervised cross-entropy loss (denoted as $\mathrm{CE}$):
\begin{equation}
    \mathcal{L}_{S} = \frac{1}{|L|} \sum_{ (\x,y) \in L} \mathrm{CE} \big( \y \;\|\; q_\theta (\x ) \big),
    \label{eqn:supervised}
\end{equation}
where $\y$ is the $K$-dimensional one-hot label constructed with $y$, and $q_\theta(\x) \in [0,1]^K$ is the output probability distribution after the softmax layer for $\x$. As for the unlabeled data, consistency regularization \cite{xie2019unsupervised,sohn2020fixmatch} over $U = \Uin \cup \Uout$ can be formulated as:
\begin{equation}
    \mathcal{L}_{CR} = \frac{1}{|U|} \sum_{\x \in U}
    \mathrm{CE} \Big(q_{\theta} \big( \mathcal{A} ( \x ),T \big) \;\|\; q_{\theta} \big( \mathcal{A}^{\prime} (\x ) \big) \Big),
    \label{eqn:cr}
\end{equation}
where $\mathcal{A}(\cdot)$ and $\mathcal{A}^{\prime}(\cdot)$ are different data augmentations. The soft target in Eq~(\ref{eqn:cr}) is usually scaled with temperature $T$:
\begin{equation}
    q_{{\theta}}^{(i)}(\x,T) = \frac{\exp{(z_i / T)}}{\sum_{j=1}^{K} \exp{(z_j / T)}},
\end{equation}
where $z_i$ is the output logit of class $i$ for $\x$. For the OOD samples in $\Uout$, $T<1$ will sharpen the output probability distribution, leading to a high probability of identifying OOD samples as ID samples, while $T>1$ will soften the output probability distribution, leading to a low probability of identifying OOD samples as ID samples. Therefore, temperature $T$ should be larger than $1$ in order to exploit the OOD samples in $\Uout$. As for the ID samples in $\Uin$, although $T<1$ was used in previous works \cite{xie2019unsupervised,sohn2020fixmatch} to encourage the high-confidence output, we still set $T>1$ in Eq~(\ref{eqn:cr}) since exploiting the OOD samples in $U$ plays an important role in OOD detection. Actually, we can provide the following Theorem~\ref{theo} to show that $T>1$ will push the output confidence of ID and OOD samples further apart from each other.

\begin{theorem}\label{theo}
    Let $\score{\x, T} = q^{(\hat{y})}_{\theta}(\x,T)$ be the output confidence of the sample $\x$ with temperature $T$, where $\hat{y}$ is the predicted class on $\x$. Under $K=2$, for the samples $\xin$ and $\xout$ that satisfy $\score{\xin, 1} > \score{\xout, 1}$: \\if $c > T_1 > T_2 \geq 1$, we have
    \begin{small}\begin{equation}
        \score{\xin, T_1} - \score{\xout, T_1} > \score{\xin, T_2} - \score{\xout, T_2};
        \label{eqn:theo1}
    \end{equation}\end{small}if $0< T_1 < T_2 \leq 1$, we have
    \begin{small}\begin{equation}
        \score{\xin, T_1} - \score{\xout, T_1} < \score{\xin, T_2} - \score{\xout, T_2}.
    \label{eqn:theo2}
    \end{equation}\end{small}Here, the constant $c$ depends on $\score{\xin, 1}$ and $\score{\xout, 1}$.
\end{theorem}

\begin{remark}
    \normalfont Previous works \cite{hendrycks2016baseline,liang2017enhancing,Hsu2020generalized} have shown that ID samples tend to have higher output confidence than OOD samples, i.e., $\score{\xin, 1} > \score{\xout, 1}$ holds for most ID and OOD samples. For these ID and OOD samples, Theorem~\ref{theo} tells that the output confidence gap between them relates to temperature $T$, and $T > 1$ will make the gap larger. Due to space constraints, we postpone the omitted proof of Theorem~\ref{theo} to Appendix~A in the supplementary material.
\end{remark} 

For $K \geq 3$, it is difficult to analyze the case since $\score{\x, T}$ depends on the exponential functions. Intuitively, similar results to Theorem~\ref{theo} can be got for $K \geq 3$. For a sample $\x$, the output probability distribution of the trained neural network is denoted as $\bm{p} = (p_1,  \cdots, p_{K})$, where $p_i = \frac{\exp{(z_i)}}{\sum_{j=1}^{K} \exp{(z_j)}}$. In deep learning, neural networks are trained on the data with one-hot labels. Because of the expressive power of neural networks, the output $\bm{p}$ is usually close to a one-hot vector. Let $\tilde{\bm{p}} = (\tilde{p}_1,  \cdots, \tilde{p}_{K})$ be the decreasingly sorted version of $\bm{p}$, where $\tilde{p}_1 > \cdots > \tilde{p}_{K}$. Since $\bm{p}$ is close to a one-hot vector, it is reasonable to assume that $\tilde{p}_1 \gg \tilde{p}_i$ for $2 \leq i \leq K$. Here, we consider a weaker assumption that $\tilde{p}_1 \gg \tilde{p}_j$ for $3 \leq j \leq K$, which can be met in realistic applications (we will verify this assumption with experiments in Appendix~C in the supplementary material). For the trained neural network $f_\theta$, we have
    \begin{align}
        \score{\x, T} &= 1 / \big(1 + \sum_{i \neq \hat{y}}\exp{(z_i / T - z_{\hat{y}} / T )} \big) \notag \\
        &= 1 / \big(1 + \sum_{j=2}^{K} (\tilde{p}_j / \tilde{p}_1)^{1 / T}\big).
    \end{align}
    For $3 \leq j \leq K$, $\tilde{p}_j / \tilde{p}_1 \approx 0$ since $\tilde{p}_1 \gg \tilde{p}_j$, so we have $\score{\x, T} \approx 1 / \big(1 + (\tilde{p}_2 / \tilde{p}_1)^{1 / T}\big)$. With similar proof of Theorem~\ref{theo}, we can obtain Eq~(\ref{eqn:theo1}) and Eq~(\ref{eqn:theo2}) under $K \geq 3$.
\makeatletter
\renewcommand{\ALG@name}{Procedure}
\makeatother

\begin{algorithm}[!b]
	\caption{Obtaining the thresholds $\thtin$ and $\thtout$}
	\label{alg:gmm}
	\textbf{Input:} Unlabeled data $U$, epoch $t$, neural network $f_\theta$.\\
	\textbf{Parameter:} Components $g_1$ (for ID) and $g_2$ (for OOD) of Gaussian Mixture Model (GMM).
	\begin{algorithmic}[1]
        \STATE Fit $g_1$ and $g_2$ on $\big\{ \score{\x, T_t} \;|\; \x \in U \big\}$ with EM;
        \STATE Separate $U$ with $g_1$ and $g_2$ by the posterior probability: \\
            \quad $U^{g_1}_t \gets \big\{ \x \;|\; p \big( g_1 | \score{\x, T_t} \big) > p \big(g_2 | \score{\x, T_t} \big) \land$ \\
            \quad $\x \in U \big\}$; \\
            \quad $U^{g_2}_t \gets \big\{ \x \;|\; p \big( g_1 | \score{\x, T_t} \big) \leq p \big(g_2 | \score{\x, T_t} \big) \land$ \\
            \quad $\x \in U \big\}$; \\
        \STATE Calculate the thresholds:\\
            \quad $\thtin \gets \frac{1}{|U^{g_1}_t|} \sum_{\x \in U^{g_1}_t} \score{\x, T_t}$; \\
            \quad $\thtout \gets \frac{1}{|U^{g_2}_t|} \sum_{\x \in U^{g_2}_t} \score{\x, T_t}$.
	\end{algorithmic}
	\textbf{Output:} $\thtin$ and $\thtout$.
\end{algorithm}

However, $T>1$ will soften the output probability distribution of the ID samples in $\Uin$. In order to guarantee the performance of ID classification, we employ the calibration technique \cite{guo2017on} to set the value of $T$ adaptively. Specifically, $T$ is optimized with respect to the negative log likelihood loss on the ID validation set $V$ in epoch $t$:
\begin{equation}
    T_t = \mathop{\arg\min}_{T} \; - \sum_{(\x,y) \in V} 
    \log\left( q^{(y)}_{\theta} (\x, T) \right).
    \label{eqn:calib}
\end{equation}
The temperature $T_t$ will push the output confidence of the ID and OOD samples in $U$ further apart from each other, which motivates us to select these samples from $U$ with the output confidence. A straightforward way is to employ the constant confidence thresholds $\thin$ and $\thout$, i.e., the sample $\x$ is determined as ID if $\score{\x, T} > \thin$ or OOD if $\score{\x, T} < \thout$. However, since the output confidence distribution of the samples in $U$ varies during the training process, we should choose the thresholds $\thin$ and $\thout$ dynamically. We fit a two-component Gaussian Mixture Model (GMM) on the output confidence of the samples in $U$ with the Expectation-Maximization algorithm in each epoch and calculate the average confidence of the samples in each component as the confidence threshold. This process is summarized in Procedure~\ref{alg:gmm}, and the selected ID and OOD samples in epoch $t$ can be written as:
\begin{align}
    \Utin &= \big\{ \x \;|\; \x \in U \land \score{\x, T_t} > \thtin \big\}, 
    \label{eqn:select_id} \\
    \Utout &= \big\{ \x \;|\; \x \in U \land \score{\x, T_t} < \thtout \big\}.
    \label{eqn:select_ood}
\end{align}
 
There may be samples of unseen OOD classes in the test data during the inference process (shown in Figure~\ref{fig:setting_explain}), so it is important to enhance the model's generalization capability to deal with them. One reasonable way is to increase the diversity of data with augmentation techniques. Here we employ RandAugment \cite{randaug} and mixup \cite{zhang2018mixup}. RandAugment can produce heavily distorted versions of a given image, and mixup aims to generate new samples and targets by linear combination. We modify the vanilla mixup since there are OOD samples in $U$. Specifically, we first choose $\lambda$ from $Beta(\alpha, \alpha)$, where $\alpha$ is a hyperparameter of the Beta distribution. Then, for a pair $(\x, \x^\prime)$, we obtain the new sample $\hat{\x} = \lambda^{\prime} \x + (1 - \lambda^{\prime}) \x^\prime$, where $\x, \x^\prime \in U$. $\lambda^\prime = \max(\lambda, 1 - \lambda)$ leads $\hat{\x}$ closer to $\x$ to preserve the semantic information in the original image $\x$. We combine these two augmentation techniques to further improve the richness of data, which can be formulated as:
\begin{align}
    \tilde{\x} &= \mathcal{M} \big( \mathcal{R} (\x) , \mathcal{R} ( \x^{\prime} ) \big) \notag \\
               &= \lambda^{\prime} \mathcal{R} (\x) + (1 - \lambda^{\prime}) \mathcal{R} (\x^{\prime}),
\end{align}
where $\mathcal{R}(\cdot)$ denotes RandAugment for a given sample.

In order to make the model distinguish between ID and OOD samples better with the output confidence, we employ entropy minimization and maximization on the augmented samples in $\Utin$ and $\Utout$ respectively. The learning objectives can be written as:
\begin{align}
    \mathcal{L}_{Emin} &= \frac{1}{|U|} \sum_{\x \in \Utin} \mathrm{CE} \big(\hat{q}_{\theta}(\x) \;\|\; q_{\theta}(\tilde{\x}) \big),
    \label{eqn:emin} \\
    \mathcal{L}_{Emax} &= - \frac{1}{|U|} \sum_{\x \in \Utout} \mathrm{H} \big( q_{\theta}(\tilde{\x}) \big).
    \label{eqn:emax}
\end{align}
In Eq~(\ref{eqn:emin}), $\hat{q}_{\theta}(\x)$ is the $K$-dimensional one-hot pseudo label for $\x$, in which the $i$th element $\hat{q}_{{\theta}}^{(i)}(\x)=1$ if and only if $i=\hat{y}$ ($\hat{y}$ is the predicted class on $\x$). It enforces the model to produce the low-entropy (high-confidence) output on ID samples. In Eq~(\ref{eqn:emax}), $\ent{\cdot}$ calculates the entropy of a given distribution and enforces the model to produce the high-entropy (low-confidence) output on OOD samples. Similar to \citet{sohn2020fixmatch}, we normalize these losses with $|U|$ to take the capacity of the selected sets into consideration.

In this way, the overall loss can be formulated as:
\begin{equation}
    \mathcal{L} = \mathcal{L}_{S} + \omega \mathcal{L}_{CR} + \beta \mathcal{L}_{Emin} + \gamma \mathcal{L}_{Emax},
    \label{eqn:overall}
\end{equation}
where $\omega$, $\beta$, and $\gamma$ are hyperparameters to balance each loss. We first obtain a basic model with $L$ and $U$, and use $\omega=1$ and $\beta=\gamma=0$ at the first stage of the training process. After getting the basic model, we select the ID and OOD samples in $U$ with the output confidence and use $\omega=0$ and $\beta=\gamma=1$. This intuition is similar to curriculum learning \cite{bengio2009curriculum}, which starts with an easier learning objective and then faces a more difficult one. The overall training process of \algfull is shown in Algorithm~\ref{alg:AIOL}, and the output confidence is used as the detection score during the inference process.

\makeatletter
\renewcommand{\ALG@name}{Algorithm}
\makeatother

\begin{algorithm}[!t]
    \setcounter{algorithm}{0}
	\caption{\algfull}
	\label{alg:AIOL}
	\textbf{Input:} Labeled data $L$, Unlabeled data $U$. \\
	\textbf{Parameter:} Neural network $f_\theta$.
	\begin{algorithmic}[1]
		\FOR {$t=1$ \textbf{to} \textit{max\_epoch}}
            \STATE Obtain temperature $T_t$ with Eq~(\ref{eqn:calib});
            \STATE Obtain thresholds $\thtin$ and $\thtout$ with Procedure~\ref{alg:gmm};
            \FOR {$k=1$ \textbf{to} \textit{max\_iteration}}
                \STATE Draw a batch of labeled data $B_L$ from $L$, and draw a batch of unlabeled data $B_U$ from $U$;
                \STATE Compute $\mathcal{L}_S$ with Eq~(\ref{eqn:supervised}) on $B_L$;
                \STATE Compute $\mathcal{L}_{CR}$, $\mathcal{L}_{Emin}$, and $\mathcal{L}_{Emax}$ with Eq~(\ref{eqn:cr}), Eq~(\ref{eqn:emin}), and Eq~(\ref{eqn:emax}) respectively on $B_U$;
                \STATE Update neural network parameters $\theta$ with Eq~(\ref{eqn:overall}).
            \ENDFOR
        \ENDFOR
	\end{algorithmic}
	\textbf{Output:} Trained neural network $f_\theta$.
\end{algorithm}

\begin{table*}[t]
\small
\centering

\begin{tabular}{l c *4{c} c *4{c}}
\toprule
$\Uin$ && \multicolumn{4}{c}{CIFAR10} && \multicolumn{4}{c}{CIFAR100} \\
\cmidrule{1-1} \cmidrule{3-6} \cmidrule{8-11}
$\Uout$ && ImageNet & SVHN & CIFAR100 & Split  && ImageNet & SVHN & CIFAR10 & Split \\
\cmidrule{1-1} \cmidrule{3-11} 
OOD type && \multicolumn{4}{c}{Seen\s Unseen} && \multicolumn{4}{c}{Seen\s Unseen} \\
\midrule
Baseline    && 66.2\s65.1 & 54.2\s66.4 & 63.9\s65.3 & 66.4\s61.4 && 70.3\s64.3 & 70.2\s64.2 & 68.1\s64.7 & 75.5\s74.3 \\
OE          && 69.6\s70.7 & 82.9\s62.3 & 63.9\s59.3 & 65.3\s59.2 && 86.5\s82.5 & 98.7\s66.8 & 70.3\s67.9 & 76.1\s74.6 \\
MCD         && 98.4\s83.7 & 97.3\s57.7 & 59.0\s51.6 & 73.5\s57.8 && 96.1\s84.4 & 99.3\s69.3 & 79.8\s70.0 & 92.2\s82.3 \\
SSD         && 73.7\s86.9 & 33.3\s\textBF{94.8} & 81.6\s93.3 & 39.1\s94.5 && 49.3\s67.8 & 29.9\s71.5 & 47.4\s79.7 & 44.9\s75.4 \\
FixMatch    && 44.6\s85.4 & 40.2\s93.2 & 76.8\s90.7 & 39.4\s91.4 && 47.3\s72.6 & 25.0\s77.1 & 60.4\s78.0 & 39.1\s73.2 \\
UASD        && 86.0\s85.7 & 88.4\s84.7 & 80.0\s86.0 & 80.4\s82.1 && 74.6\s72.9 & 77.1\s71.3 & 71.3\s72.2 & 74.6\s73.6 \\
\midrule
Ours        && \textBF{99.9}\s\textBF{96.9} & \textBF{100.}\s93.4 & \textBF{94.7}\s\textBF{95.0} & \textBF{93.8}\s\textBF{94.6} && \textBF{99.7}\s\textBF{90.0} & \textBF{100.}\s\textBF{77.4} & \textBF{80.0}\s\textBF{80.9} & \textBF{97.7}\s\textBF{88.9}   \\
\bottomrule
\end{tabular}

\caption{OOD detection results with percentage of AUROC.}
\label{tab:main_results}
\end{table*}

\section{Experiments}

\subsection{Datasets}
The training set of CIFAR10 (CIFAR100) is used as ID training data, and we split them into labeled and unlabeled data. For the labeled data, the number of each class is set as 100, which results in 1000 (10000) labeled data for CIFAR10 (CIFAR100). The remaining data is used as the unlabeled ID data $\Uin$. As for $\Uout$, the unlabeled OOD data are drawn from the following datasets:
\begin{itemize}
    \item \textbf{ImageNet.} The test set of ImageNet, i.e., 50000 images with 1000 classes, are drawn to construct $\Uout$.
    
    \item \textbf{SVHN.} The test set of SVHN, i.e., 26032 images with 10 classes, are drawn to construct $\Uout$.
    
    \item \textbf{CIFAR10 $\&$ CIFAR100.} If CIFAR10 is considered as the ID dataset, the test set of CIFAR100, i.e., 10000 images with 100 classes, will be drawn to construct $\Uout$; if CIFAR100 is considered as the ID dataset, the test set of CIFAR10, i.e., 10000 images with 10 classes, will be drawn to construct $\Uout$.
    
    \item \textbf{Split.} Except for the mixtures of two different datasets, we also consider the splits of one single dataset. Specifically, we split CIFAR10 into \textit{animal} group as ID and \textit{non-animal} group as OOD, i.e., 6 classes and 4 classes. Similarly, we split CIFAR100 into \textit{living} group as ID and \textit{non-living} group as OOD, i.e., 65 classes and 35 classes. Details of the splits are given in Appendix~B in the supplementary material. We also keep 100 labeled data per class for the ID group and use the remaining data of the training set as the mixed unlabeled data $U$ because they naturally include the ID and OOD groups.
\end{itemize}

The test set of CIFAR10 (CIFAR100) is used to evaluate the performance. Following that in \citet{Yu2019mcd} and \citet{sastry2020gram}, we split 10\% of the test set as the ID validation data and use the rest as the ID test data. As for the OOD test data, samples of seen and unseen OOD classes are drawn as follows:
\begin{itemize}
    \item \textbf{Seen.} If $\Uin$ and $\Uout$ are from two different datasets, samples of seen OOD classes will be drawn from the training set of the dataset for $\Uout$. For example, when $\Uin$ is from CIFAR10 and $\Uout$ is from CIFAR100, samples of seen OOD classes are drawn from the training set of CIFAR100; if $\Uin$ and $\Uout$ are from the splits of one single dataset, samples of seen OOD classes will be drawn from the OOD group in the test set. For example, when $\Uin$ and $\Uout$ are from the splits of CIFAR10, samples of seen OOD classes are drawn from the \textit{non-animal} group in the test set of CIFAR10.
    
    \item \textbf{Unseen.} Samples of unseen OOD classes are drawn from various benchmark datasets: CIFAR10, CIFAR100, SVHN, ImageNet, Blobs, Texture, iSUN, LSUN, and Places365. The dataset will not be used to construct the unseen OOD data if it is considered for training. Following that in \citet{hendrycks2016baseline}, each benchmark dataset is used to evaluate the detection performance respectively, and the average result is reported.
\end{itemize}

More details of the used datasets are given in Appendix~B in the supplementary material.

\subsection{Setup}
Following that in \citet{sohn2020fixmatch}, the augmentation $\mathcal{A}(\cdot)$ is implemented with the standard data augmentations (random flip and crop), and the augmentation $\mathcal{A}^{\prime}(\cdot)$ is implemented with RandAugment. In the experiments, we use the standard Wide ResNet \cite{Zagoruyko2016wide}, i.e., WRN-28-2, as the base network and use SGD optimizer for training. The experiments are run for 256 epochs with 512 iterations per epoch. We set $\omega=1$ and $\beta=\gamma=0$ at the beginning, and set $\omega=0$ and $\beta=\gamma=1$ after 80\% of the training epochs. We set $\alpha=0.2$ for mixup. We employ EMA model  \cite{tarvainen2017mean}, and limit $\thtin \leq 0.95$ and $\thtout \geq 1 / K + 0.05$ for stability. Other hyperparameters are the same as that of \citet{sohn2020fixmatch} for a fair comparison. We evaluate the detection performance with three metrics: AUROC, AUPR, and FPR95. Due to space constraints, we present results with AUROC (the area under the ROC curve) in this section. More results with other metrics are in Appendix~C in the supplementary material.

\subsection{Compared Methods}
We compare our method with various related methods, including: Baseline \cite{hendrycks2016baseline}; OE \cite{hendrycks2019deep}; MCD \cite{Yu2019mcd}; SSD \cite{Sehwag2021ssd}; FixMatch \cite{sohn2020fixmatch}; UASD \cite{Chen2020uasd}. The Baseline method only uses labeled data for training, but the others consider unlabeled data. Note that the first four methods are developed for OOD detection, while the last two methods are barely SSL methods. The OOD detector based on the output confidence is built for the last two methods, which is the same as ours. For a fair comparison, the experiments are also conducted for 256 epochs with 512 iterations per epoch on WRN28-2 for all the compared methods except for SSD. For SSD, it is a self-supervised learning method and needs more training resources, so we follow the original paper and run it on ResNet-18 \cite{he2016deep} for 500 epochs. The hyperparameters are set according to the original paper for all the compared methods.

\subsection{Results}
Since the test data contain samples of seen and unseen OOD classes, we evaluate the detection performance on them respectively, and the results are shown in Table~\ref{tab:main_results}.

\textbf{Performance of seen OOD detection.} The results for detecting samples of seen OOD classes are summarized on the left of the slash in Table~\ref{tab:main_results}, which indicates that our method outperforms the compared methods on seen OOD detection. The MCD method performs worse than ours since it treats each unlabeled sample equally. The SSD method and the FixMatch method are developed for the pure unlabeled ID data, while the OE method is developed for the pure unlabeled OOD data. These three methods all produce poor and unstable results since they are confused with the mixed ID and OOD samples in $U$. The UASD method produces more stable results since it eliminates the effect of OOD samples in $U$, but it still performs worse than ours since we can learn from OOD samples in $U$ rather than ignoring them.

\textbf{Performance of unseen OOD detection.} The results for detecting samples of unseen OOD classes are summarized on the right of the slash in Table~\ref{tab:main_results}. Except when $\Uin$ is from CIFAR10 and $\Uout$ is from SVHN, our method outperforms the compared methods because the augmentation techniques enhance the model's generalization capability on unseen OOD detection. When $\Uin$ is from CIFAR10 and $\Uout$ is from SVHN, our method performs slightly worse than the SSD method. The reason is that the samples in SVHN are images of house numbers, and it is hard to get more diverse information from these plain images. Furthermore, the SSD method requires more computational resources, which is run for 500 epochs on ResNet-18.

\begin{table}[!t]
\small
\centering

\begin{tabular}{l c *3{c}}
\toprule
$|L|$ && 250 & 1000 & 4000 \\
\cmidrule{1-1} \cmidrule{3-5}
OOD type    &&  \multicolumn{3}{c}{Seen\s Unseen} \\
\midrule
Baseline    && 53.3\s52.5 & 66.2\s65.1 & 76.7\s76.8 \\
FixMatch    && 42.1\s84.6 & 44.6\s85.4 & 68.2\s88.5 \\
UASD        && 70.4\s70.0 & 86.0\s85.7 & 90.5\s91.1 \\
\midrule
Ours        && \textBF{99.8}\s\textBF{96.1} & \textBF{99.9}\s\textBF{96.9} & \textBF{99.3}\s\textBF{97.0} \\
\bottomrule
\end{tabular}

\caption{Results with different numbers of labeled samples.}
\label{tab:labelnums}
\end{table}
\begin{figure}[!t]
    \centering
    \includegraphics[width=0.45\textwidth]{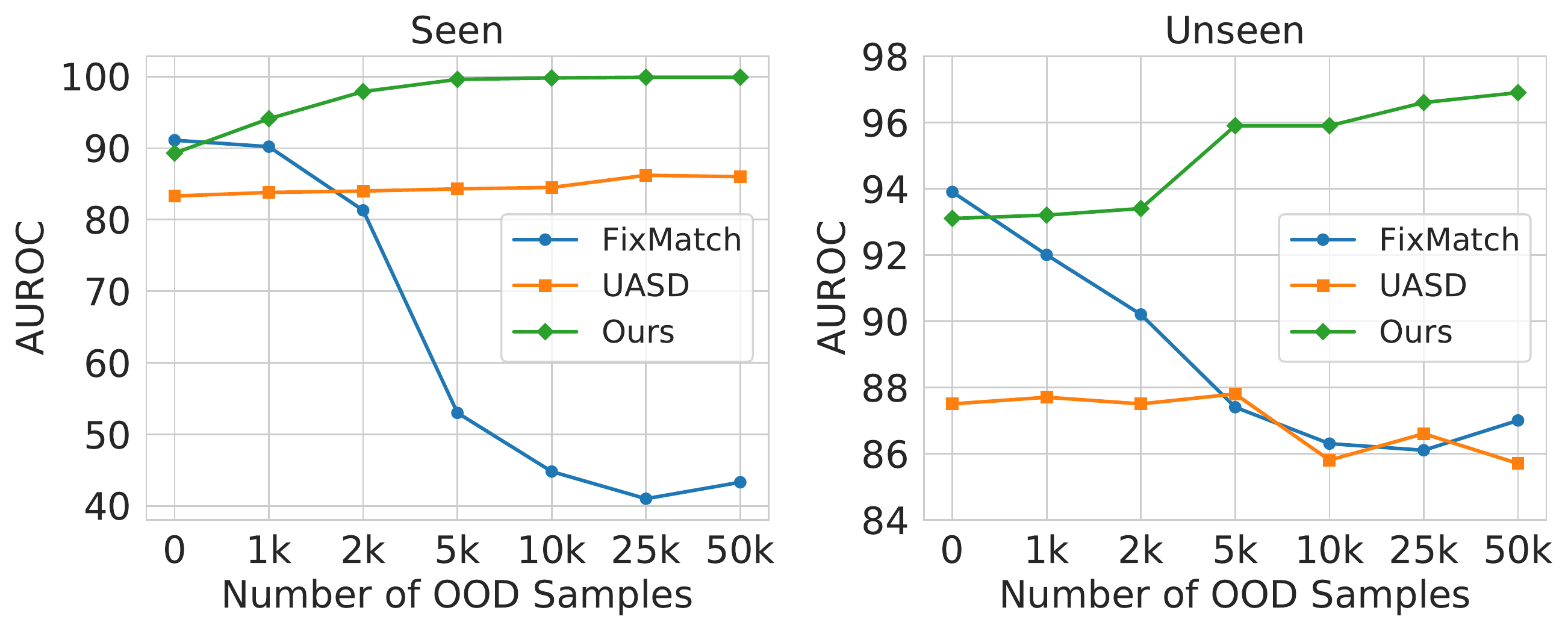}
    \caption{Results with various numbers of OOD samples.}
    \label{fig:oodnums}
\end{figure}

Since our method is developed for OOD detection with the labeled data $L$ and the mixed unlabeled data $U$, number of labeled samples and number of OOD samples in $U$ are parameters of the experiments. We set that $\Uin$ is from CIFAR10 and $\Uout$ is from ImageNet, and provide further results to investigate the effectiveness of the two parameters:

\textbf{Number of labeled samples.} We conduct the experiments with different numbers of labeled samples (250, 1000, and 4000), and the results are shown in Table~\ref{tab:labelnums}. It can be found that our method performs better than the other methods with different numbers of labeled samples and is not sensitive to this parameter. Moreover, our method can achieve superior performance even with very few labeled samples, i.e., 250. The reason lies in that our method can make full use of the mixed unlabeled data.

\textbf{Number of OOD samples in $U$.} We conduct the experiments with various numbers of OOD samples in $U$ ($|\Uout|$ varies from 0 to 50000), and the results are shown in Figure~\ref{fig:oodnums}. We set $\gamma=0$ when there is no OOD sample in $U$. Figure~\ref{fig:oodnums} shows that our method outperforms the other methods with various numbers of OOD samples in $U$ except for $|\Uout|=0$. When $|\Uout|=0$, our method is slightly worse than the FixMatch method, which is developed for the pure unlabeled ID data ($|\Uout|=0$). But our method is appropriate for the mixed unlabeled data ($|\Uout|>0$) and can produce better results with more OOD samples in $U$.

\begin{table}[t]
\small
\centering
\begin{tabular}{l c}
\toprule
Ablation & \;\;\;\;Seen\s Unseen  \\
\midrule
\;\;\; Supervised loss \big(Eq(\ref{eqn:supervised})\big) & 75.5\s74.3 \\
\makecell[l]{$+$ Consistency regularization \big(Eq~(\ref{eqn:cr})\big) \\ \;\;\; with $T=1$} & 59.2\s80.4 \\
\makecell[l]{$+$ Adaptive temperature $T_t$} & 89.7\s84.3 \\
\makecell[l]{$+$ Emin and Emax \big(Eq~(\ref{eqn:emin}) and Eq~(\ref{eqn:emax})\big) \\ \;\;\; without augmentation} & 96.6\s83.6 \\
\makecell[l]{$+$ RandAugment} & 94.0\s85.6 \\
\makecell[l]{$+$ Modified mixup} & \textBF{97.7}\s\textBF{88.9}\\
\bottomrule
\end{tabular}

\caption{Ablation study on the used modules.}
\label{tab:main_ablation}
\end{table}

\subsection{Ablation Study}
We set that $\Uin$ and $\Uout$ are from the splits of CIFAR100 and run the experiments to study the details of our method and provide additional insight into what makes it successful.

\textbf{The effectiveness of the used modules.} There are several modules in our method: supervised loss \big(Eq~(\ref{eqn:supervised})\big); consistency regularization \big(Eq~(\ref{eqn:cr})\big) with $T=1$; adaptive temperature $T_t$; Emin and Emax \big(Eq~(\ref{eqn:emin}) and Eq~(\ref{eqn:emax})\big) without augmentation; two augmentation techniques. The ablation study on these modules is summarized in Table~\ref{tab:main_ablation}. We can see that the method with supervised loss and consistency regularization ($T=1$) performs badly, but after using the adaptive temperature $T_t$, the method performs much better. Combined with Eq~(\ref{eqn:emin}) and Eq~(\ref{eqn:emax}) over $\Utin$ and $\Utout$, which are selected according to the adaptive temperature $T_t$ and the dynamic thresholds $\thtin$ and $\thtout$, the method achieves significantly better results on seen OOD detection. Further combined with the two augmentation techniques, i.e., RandAugment and the modified mixup, the model's generalization capability on unseen OOD detection is improved.

\begin{figure}[t]
    \centering
    \includegraphics[width=0.45\textwidth]{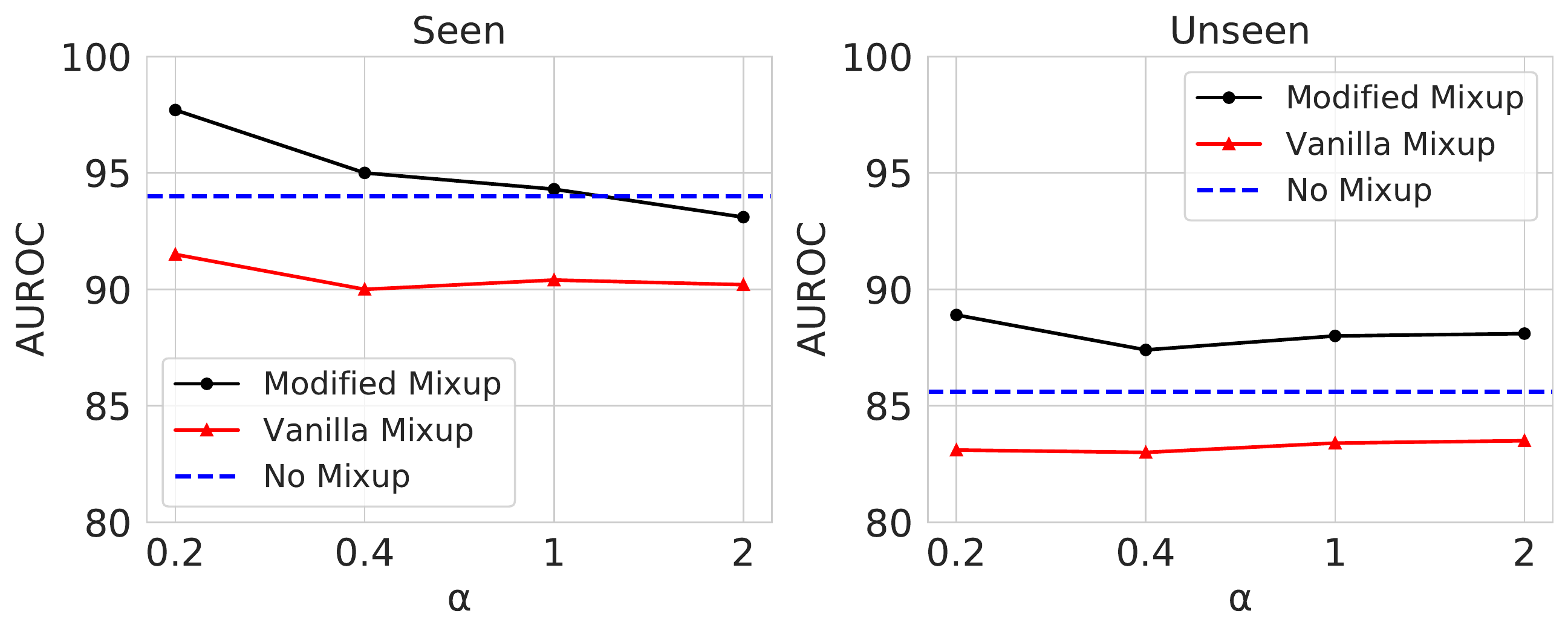}
    \caption{Ablation study on the modified mixup.}
    \label{fig:mixup}
\end{figure}
\begin{figure*}[t]
    \centering
    \subfigure[]{
        \label{fig:ccr}
        \includegraphics[width=0.28\textwidth]{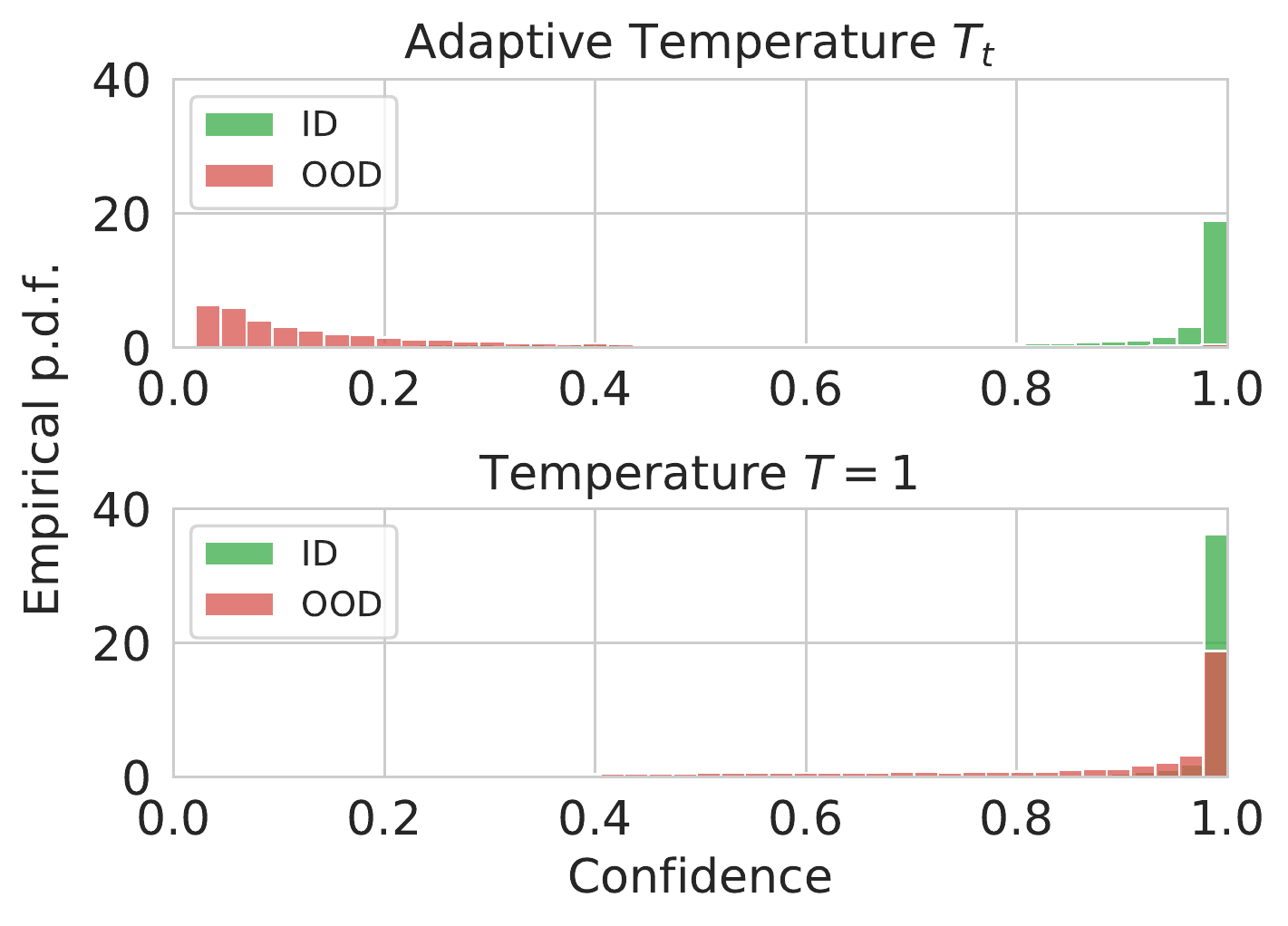}
    }\hspace{5mm}
    \subfigure[]{
        \label{fig:calib}
        \includegraphics[width=0.28\textwidth]{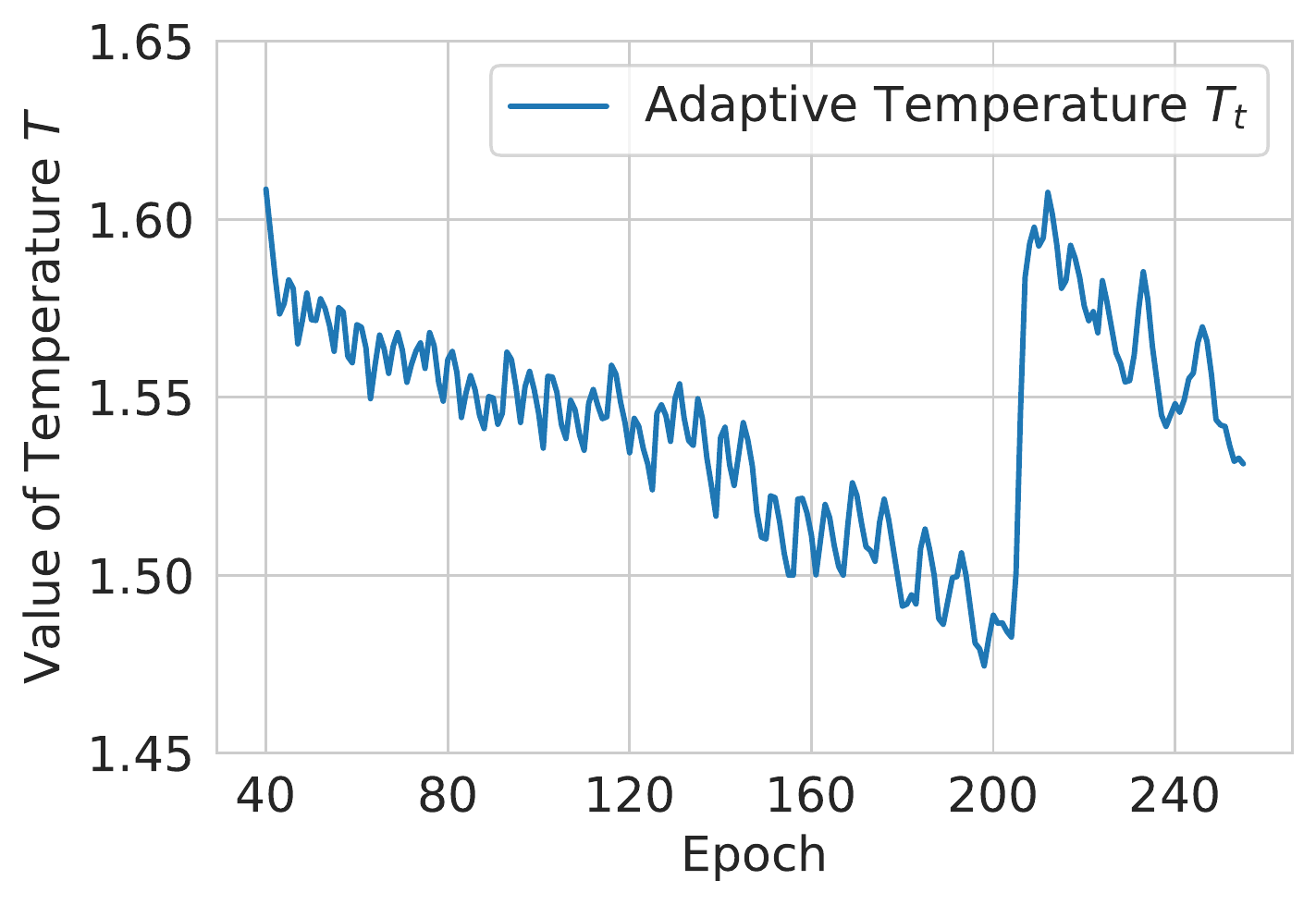}
    }\hspace{5mm}
    \subfigure[]{
        \label{fig:temperature}
        \includegraphics[width=0.28\textwidth]{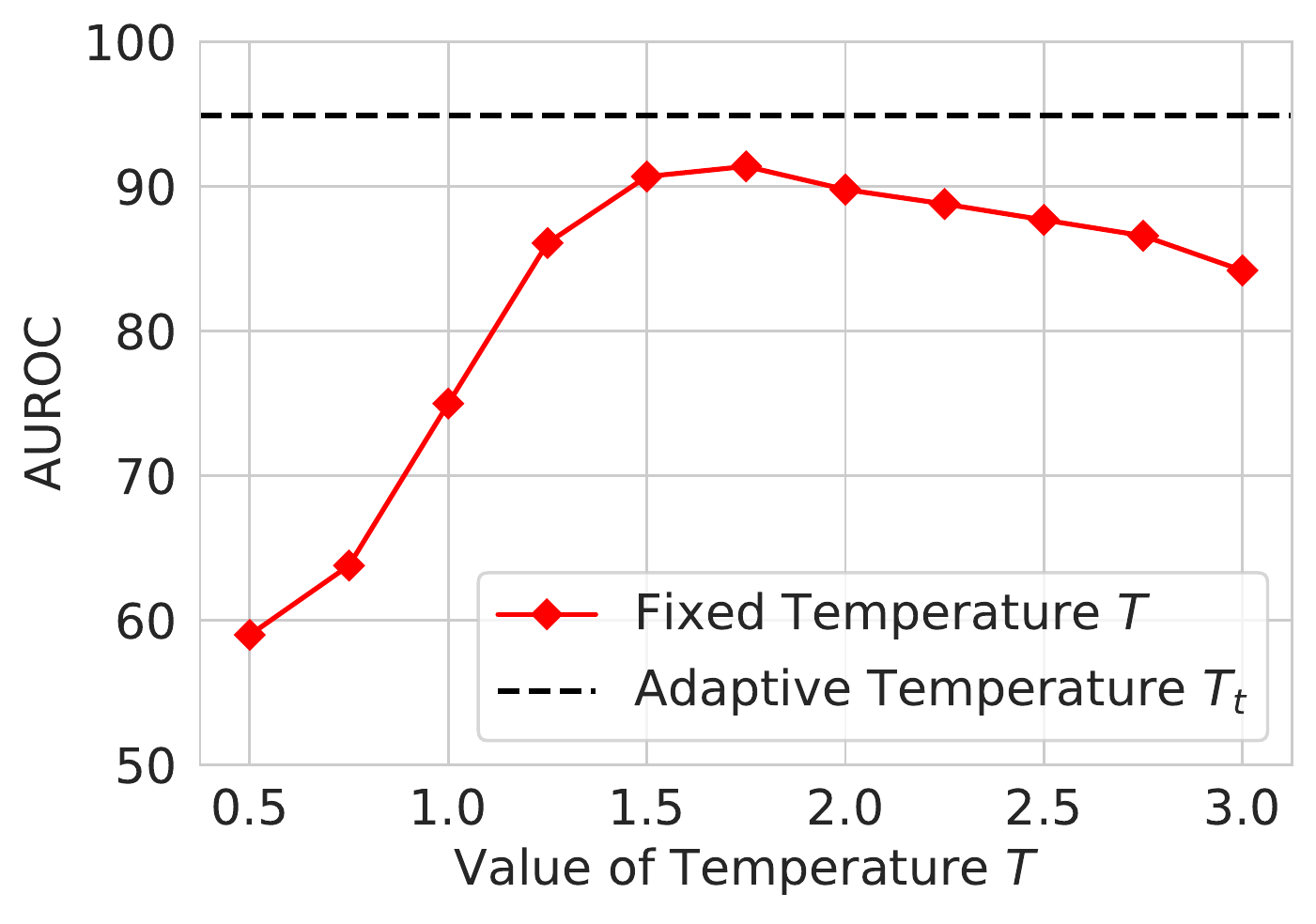}
    }
    \caption{Ablation study on the adaptive temperature $T_t$. (a): Empirical p.d.f. of the output confidence of the samples in $U$ with $T_t$ (upper part) and $T=1$ (lower part). (b): The values of $T_t$ during training. (c) Detection results on $U$ with different $T$.}
    \label{fig:cr_details}
\end{figure*}
\begin{figure}[t]
    \centering
    \subfigure[]{
        \label{fig:threshold1}
        \includegraphics[width=0.45\textwidth]{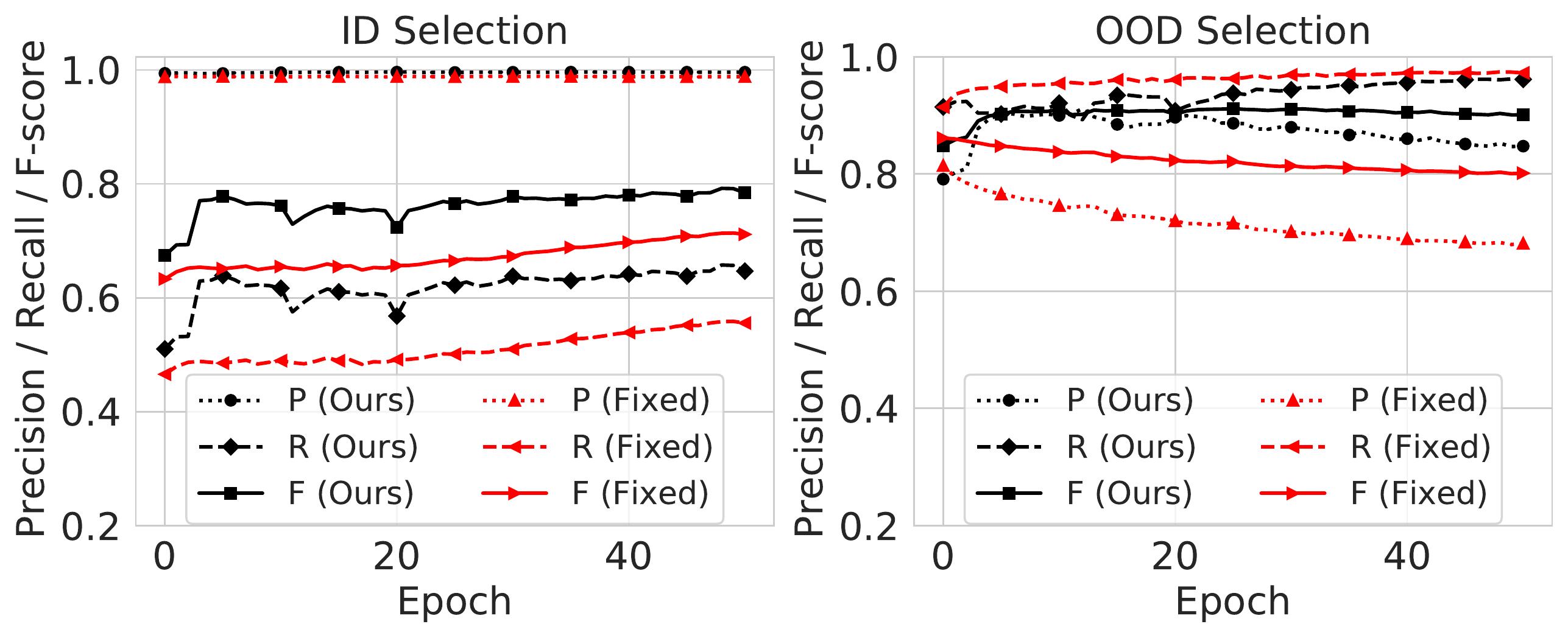}
    }
    \subfigure[]{
        \label{fig:threshold2}
        \includegraphics[width=0.45\textwidth]{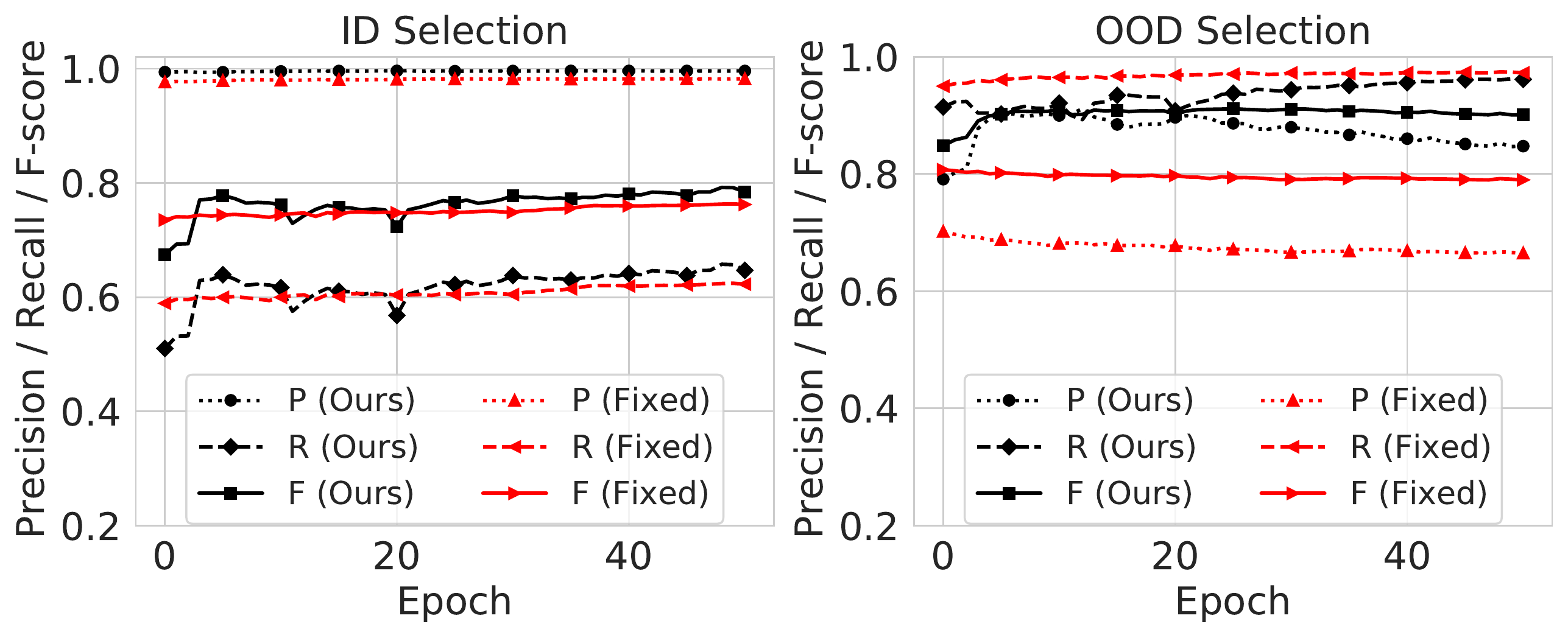}
    }
    \caption{Precision (P), Recall (R) and F-score (F) of the selection for ID samples (left part) and OOD samples (right part) with our dynamic thresholds ($\thtin$ and $\thtout$) and the fixed thresholds ($\thin$ and $\thout$). (a): $\thin=0.9$ and $\thout=0.3$. (b): $\thin=0.7$ and $\thout=0.5$.}
    \label{fig:threshold}
\end{figure}

\textbf{The effectiveness of the modified mixup.} The results for the mixup with different parameters are shown in Figure~\ref{fig:mixup}. On the one hand, the method with the vanilla mixup (without the maximum operation $\lambda^\prime = \max(\lambda, 1 - \lambda)$) performs worse than that with no mixup since the vanilla mixup directly confuses the ID and OOD samples. On the other hand, our method with the modified mixup performs better since $\lambda^\prime$ is close to 1 and can produce diverse data points around the original one, which can protect the semantic information. As for the hyperparameter $\alpha$, the smaller one can achieve better results since smaller $\alpha$ leads $\lambda$ closer to $1$ and bigger $\alpha$ leads $\lambda$ closer to $0.5$.

\textbf{The effectiveness of the adaptive temperature $T_t$.} After 80\% of the training epochs, we depict the output confidence of the samples in $U$ with the adaptive temperature $T_t$ in the upper part of Figure~\ref{fig:ccr}. As a comparison, we use $T=1$ and depict the output confidence of the samples in $U$ in the lower part of Figure~\ref{fig:ccr}. It can be found that the adaptive temperature $T_t$ pushes the output confidence of the ID and OOD samples further apart from each other. The values of the adaptive temperature $T_t$ during the training process are shown in Figure~\ref{fig:calib} (we use $T=1$ before epoch 40 for warming up), which indicates that $T_t$ is always larger than $1$. It rapidly increases at 80\% of the training epochs, i.e., epoch 205, because the large temperature is required to calibrate the output confidence of ID samples for Eq~(\ref{eqn:emin}). For comparison, we also run the experiments with the fixed temperature $T$, and the detection performance on $U$ is shown in Figure~\ref{fig:temperature}. It demonstrates that the adaptive temperature $T_t$ can consistently outperform the fixed ones.

\textbf{The effectiveness of the thresholds $\thtin$ and $\thtout$.} For obtaining $\Utin$ in Eq~(\ref{eqn:select_id}) and $\Utout$ in Eq~(\ref{eqn:select_ood}) from $U$, we choose the dynamic thresholds $\thtin$ and $\thtout$ based on the two GMM components $g_1$ and $g_2$. For comparison, we also run the experiments with the fixed thresholds ($\thin=0.9$ and $\thout=0.3$, or $\thin=0.7$ and $\thout=0.5$), and report the Precision, Recall, and F-score of the selection on $U$ in Figure~\ref{fig:threshold1} and Figure~\ref{fig:threshold2}. F-score, i.e., the harmonic mean of Precision and Recall, is a metric that can generally evaluate the selection quality. From Figure~\ref{fig:threshold1} and Figure~\ref{fig:threshold2}, we can see that the F-score of our dynamic thresholds is always higher than that of the fixed thresholds. Detailed results for more fixed thresholds are given in Appendix~C in the supplementary material. 

\begin{table}[!t]
\small
\centering

\resizebox{\columnwidth}{!}{
\begin{tabular}{l c *4{c}}
\toprule
$\Uout$ && ImageNet & SVHN & CIFAR & Split \\
\cmidrule{1-1} \cmidrule{3-6}
$\Uin$    &&  \multicolumn{4}{c}{CIFAR10\s CIFAR100} \\
\midrule
Baseline    && 58.1\s57.1 & 58.1\s57.1 & 58.1\s57.1 & 51.6\s53.3\\
FixMatch    && 93.3\s67.9 & 94.4\s69.3 & 93.9\s70.4 & 93.6\s67.1\\
UASD        && 84.7\s64.9 & 85.1\s65.3 & 85.8\s65.8 & 82.5\s62.7\\
\midrule
Ours        && 92.6\s68.4 & 93.3\s67.7 & 93.0\s67.8 & 93.8\s66.0\\
\bottomrule
\end{tabular}
}
\caption{Classification results with percentage of accuracy.}
\vspace{-0.9mm}
\label{tab:acc}
\end{table}

\textbf{Performance of ID classification.} The results of ID classification are reported in Table~\ref{tab:acc}. Our method performs slightly worse than the FixMatch method, perhaps because we use temperature $T>1$ for the samples of $\Uin$ in Eq~(\ref{eqn:cr}). But our method is developed for detecting OOD samples and performs much better than the FixMatch method in OOD detection, which is shown in Table~\ref{tab:main_results}.

\section{Conclusion}
In this paper, we focus on the more realistic OOD detection scenario, where limited labeled data and abundant mixed unlabeled data are available for training. During the inference process, the trained model should not only detect samples of seen OOD classes but also detect samples of unseen OOD classes. We propose the \algfull method, in which we adaptively select potential ID and OOD samples from the mixed unlabeled data and optimize the entropy over them. Moreover, data augmentation techniques are brought into the method to further improve the performance of unseen OOD detection. The experimental results show that our method outperforms the compared methods on various benchmark datasets.

\section{Acknowledgments}
This work is supported by the National Key Research and Development Program of China (2018AAA0101100), the National Science Foundation of China (61921006, 61673202), and the Collaborative Innovation Center of Novel Software Technology and Industrialization.

\bibliography{aaai22}

\end{document}